\documentclass[conference]{IEEEtran}  

\IEEEoverridecommandlockouts                              

\usepackage{cite}
\usepackage{amsmath,amssymb,amsfonts}
\usepackage{graphicx}
\usepackage{textcomp}
\usepackage{xcolor}
\usepackage{booktabs}
\usepackage{hyperref}
\usepackage{algorithm}
\usepackage{algpseudocode}
\usepackage{caption}
\title{\LARGE \bf
STREAMS: An Assistive Multimodal AI Framework for Empowering Biosignal Based Robotic Controls}

\author{\IEEEauthorblockN{1\textsuperscript{st} Ali Rabiee}
\IEEEauthorblockA{\textit{Dept. of Electrical Engineering} \\
\textit{University of Rhode Island}\\
Kingston, RI, USA \\
ali.rabiee@uri.edu}
~\\
\and
\IEEEauthorblockN{2\textsuperscript{nd} Sima Ghafoori}
\IEEEauthorblockA{\textit{Dept. of Electrical Engineering} \\
\textit{University of Rhode Island}\\
Kingston, RI, USA \\
sima.ghafoori@uri.edu}

~\\
\and
\IEEEauthorblockN{3\textsuperscript{rd} Xiangyu Bai}
\IEEEauthorblockA{\textit{Dept. of Electrical and Computer Engineering} \\
\textit{Northeastern University}\\
Boston, Massachusetts, USA \\
bai.xiang@northeastern.edu}
~\\
\and
\IEEEauthorblockN{4\textsuperscript{th} MH Farhadi}
\IEEEauthorblockA{\textit{Dept. of Electrical Engineering} \\
\textit{University of Rhode Island}\\
Kingston, RI, USA \\
mh.farhadi@uri.edu}

\and
\IEEEauthorblockN{5\textsuperscript{th} Sarah Ostadabbas}
\IEEEauthorblockA{\textit{Dept. of Electrical and Computer Engineering} \\
\textit{Northeastern University}\\
Boston, Massachusetts, USA \\
ostadabbas@ece.neu.edu}

\and
\IEEEauthorblockN{6\textsuperscript{th} Reza Abiri}
\IEEEauthorblockA{\textit{Dept. of Electrical Engineering} \\
\textit{University of Rhode Island}\\
Kingston, RI, USA \\
reza\_abiri@uri.edu}
*Corresponding author
}
\begin{document}

\maketitle

\begin{abstract}
End-effector based assistive robots face persistent challenges in generating smooth and robust trajectories when controlled by human's noisy and unreliable biosignals such as muscle activities and brainwaves. The produced endpoint trajectories are often jerky and imprecise to perform complex tasks such as stable robotic grasping. We propose STREAMS (Self-Training Robotic End-to-end Adaptive Multimodal Shared autonomy) as a novel framework leveraged deep reinforcement learning to tackle this challenge in biosignal based robotic control systems. STREAMS blends environmental information and user input into a Deep Q Learning Network (DQN) pipeline for an interactive end-to-end and self-training mechanism to produce smooth trajectories for the control of end-effector based robots. The proposed framework achieved a high-performance record of 98\% in simulation with dynamic target estimation and acquisition without any pre-existing datasets. As a zero-shot sim-to-real user study with five participants controlling a physical robotic arm with noisy head movements, STREAMS (as an assistive mode) demonstrated significant improvements in trajectory stabilization, user satisfaction, and task performance reported as a success rate of 83\% compared to manual mode which was 44\% without any task support. STREAMS seeks to improve biosignal based assistive robotic controls by offering an interactive, end-to-end solution that stabilizes end-effector trajectories, enhancing task performance and accuracy. The STREAMS codes and demo videos can be accessed at: \href{https://github.com/AbiriLab/STREAMS}{https://github.com/AbiriLab/STREAMS}
\end{abstract}
\section{INTRODUCTION}
Assistive technologies aim to restore independence for individuals with severe motor impairments by providing various control interfaces, including non-invasive options like head arrays \cite{jackowski2017head}, body-machine interfaces \cite{rizzoglio2023non, lee2024learning}, eye-tracking systems \cite{wohle2021towards, admoni2016predicting}, electroencephalogram (EEG) \cite{rabiee2024wavelet, abiri2019comprehensive, ghafoori2024bispectrum}, and electromyography (EMG) muscle activity \cite{teramae2017emg}, as well as invasive solutions such as cortical implants \cite{ehrlich2022adaptive}. 
\begin{figure}[ht!] 
\centering
\includegraphics[width=3.3in]{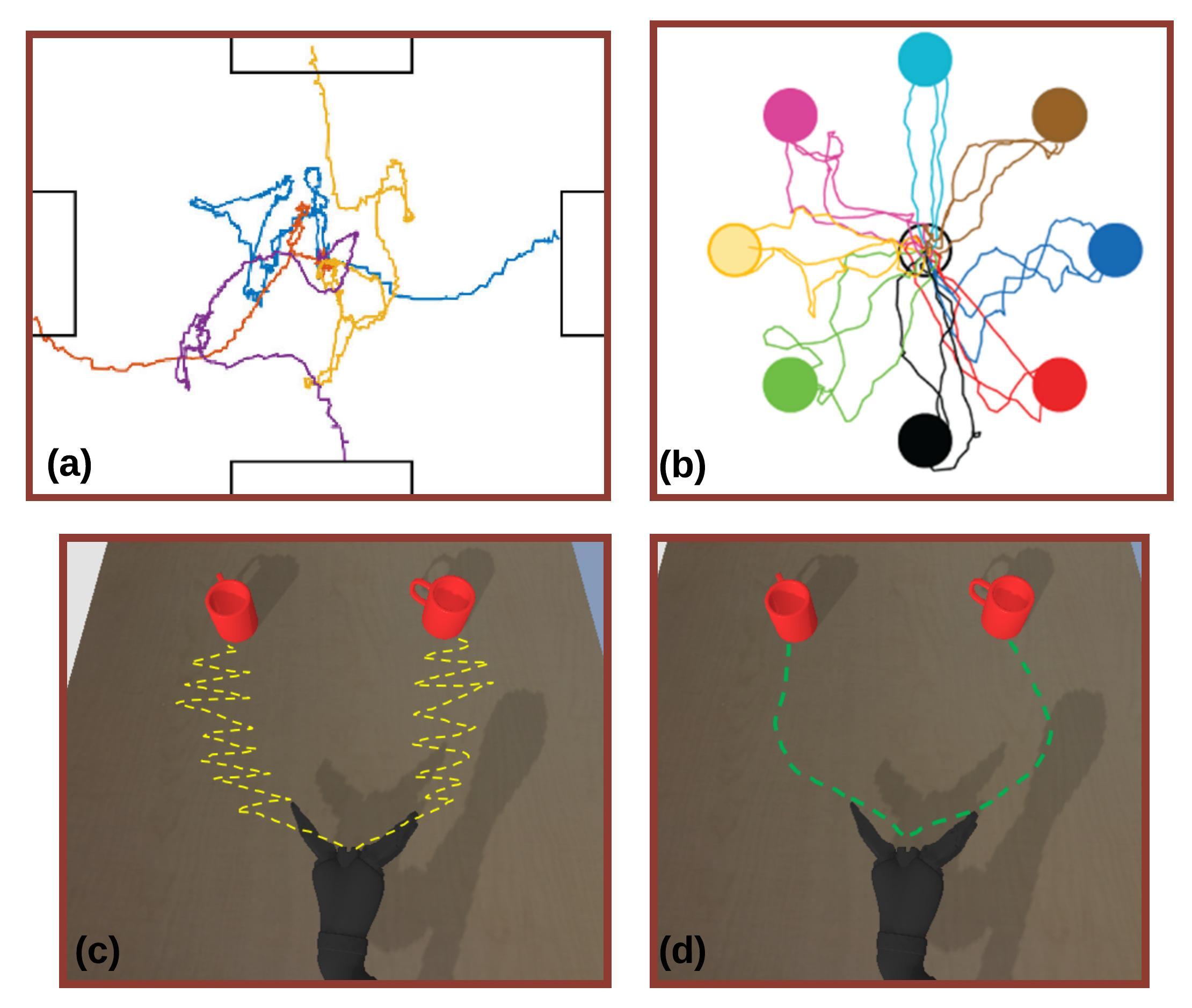}
\caption{Example of traveled trajectories by the end-effector based systems. (a) EEG-based \cite{abiri2020usability} and (b) Electrocorticography (ECoG)-based brain computer interface (BCI) studies \cite{silversmith2021plug} for computer cursor control tasks (adopted from our previous studies). The projection of such generated jerky trajectories in controlling the endpoint of robotic arms showed low performance in success rate \cite{kilmarx2018sequence}. (c) Conceptual representation of robotic arm trajectories with jerky and unreliable human input. (d) Conceptual representation of stable robotic arm trajectories modified by assistive algorithms such as STREAMS, demonstrating smoother paths to targets.}
\label{cursor_vs_robot}
\end{figure}
Among these methods, head movement-based control has gained increasing attention due to its non-invasiveness, low cost, and ease of use. Head movement signals are typically captured using inertial measurement units (IMUs), gyroscopes, or optical tracking, and are then processed by discrete-state decoders to produce a limited set of commands, such as turning left or right, moving forward, or stopping  \cite{jackowski2017head}. However, such individualized biosignal based control approaches and their associated decoders are not able to generate stable outputs (trajectories) particularly when facing the challenge of continuous control of an end-effector based system (e.g. computer cursor) for accurate target acquisition (Fig \ref{cursor_vs_robot})\cite{abiri2020usability, silversmith2021plug}. This challenge could be more destructive when controlling the endpoint of a physical robotic system where a minimum jerky motion causes failure in tasks such as robotic grasping for a patient with severe motor impairment \cite{billard2019trends}. 
Additionally, the required high level of engagement for continuous control of traveled trajectory imposes a high cognitive load on most patients particularly those with brain implants \cite{silversmith2021plug}. This cognitive burden was further escalated by the uncertainty of human intention decoders in real-world scenarios of human-robot collaboration for target acquisition, as studied by Dani et al. \cite{dani2020human}. Most of the previous research studies on shared human-robot control \cite{dragan2013policy, gopinath2016human, carlson2012collaborative, gualtieri2017open} assumed known or explicitly predicted targets, lacking real-time responsiveness to dynamic user intentions, demands, or changes in the environment's structure. To this end, we present STREAMS (Self-Training Robotic End-to-end Adaptive Multimodal Shared autonomy), a novel framework using deep reinforcement learning along with a DQN approach, to overcome the barriers of current human-robot shared control approaches in reach-to-grasp objects in 2D workspace by focusing on the following highlighted contributions:
\begin{itemize}
\item Proposed an adaptive trajectory generation method using deep reinforcement learning to produce smooth, reliable paths in complex, dynamic environments, even under noisy and imprecise biosignal control inputs. 
\item Developed an interactive end-to-end multimodal framework that directly translates unreliable biosignal inputs and environmental perception into appropriate robotic actions. This approach eliminates the need for intermediate representations or explicit intention recognition.
\item Designed a self-training DQN-based framework that eliminates the need for any datasets by leveraging synthetic data that emulates real-world biosignal noise.
\item Demonstrated a zero-shot sim-to-real translation of our framework and validated the generalizability and adaptivity of our pipeline for practical cases through a user study.
\end{itemize}

\section{Related work}
Shared autonomy paradigms in end-effector based assistive robots have explored different approaches such as sequential, arbitrated, and learning policies to balance user control and robotic assistance. Jain et al. \cite{jain2015assistive} introduced a shared control paradigm blending user inputs with the perception of the environment through sequential policies. However, their system demanded significant user input and imposed a high cognitive load. Additionally, it relied on pre-programmed actions, limiting its adaptability to new conditions and diverse objects. Downey et al. \cite{downey2016blending} developed a shared control paradigm that arbitrated between a neural spikes decoder and autonomous robotic assistance for object manipulation. Their approach was constrained by pre-defined rules without interactivity with users, making it unable to dynamically adapt to user inputs or evolving situations. Muelling et al. \cite{muelling2017autonomy} developed a continuous assistance method using arbitrated policies based on a probability distribution over goals but was not able to address the challenge of handling imprecise user inputs and varied intents. Xu et al. \cite{xu2019shared} reported a shared control paradigm that arbitrated between an EEG-based brain-computer interface (BCI) and computer vision guidance based on the robotic arm's endpoint location relative to the target. However, it was tested only with fixed or randomly placed single targets and did not address more complex scenarios or multi-object environments. Beraldo et al. \cite{beraldo2022shared} presented a shared autonomy paradigm that fused user input and environmental context by relying on a set of predefined policies to predict the next subgoal for a navigation task. While effective for navigation, the system's ability to generalize to more complex tasks, such as reach-to-grasp, was not evaluated. Additionally, the reliance on predefined policies restricts the adaptability of the system to dynamically changing environments and user preferences. Furthermore, a key limitation of most methods for end-effector based robots is their dependency on user datasets for training, explicit goal prediction, and predefined policies. In contrast, our work presented an end-to-end self-training approach that delivers continuous and interactive assistance in dynamic multi-target environments in which the user intention may be ambiguous without depending on datasets. It could be adaptable to a wide range of scenarios and prioritizes stabilizing endpoint trajectories, enabling precise robotic grasping in assistive robotics. 

\section{Methods}

\subsection{Problem formulation and framework}

\begin{figure*}[ht!] 
\centering
\includegraphics[width=7in]{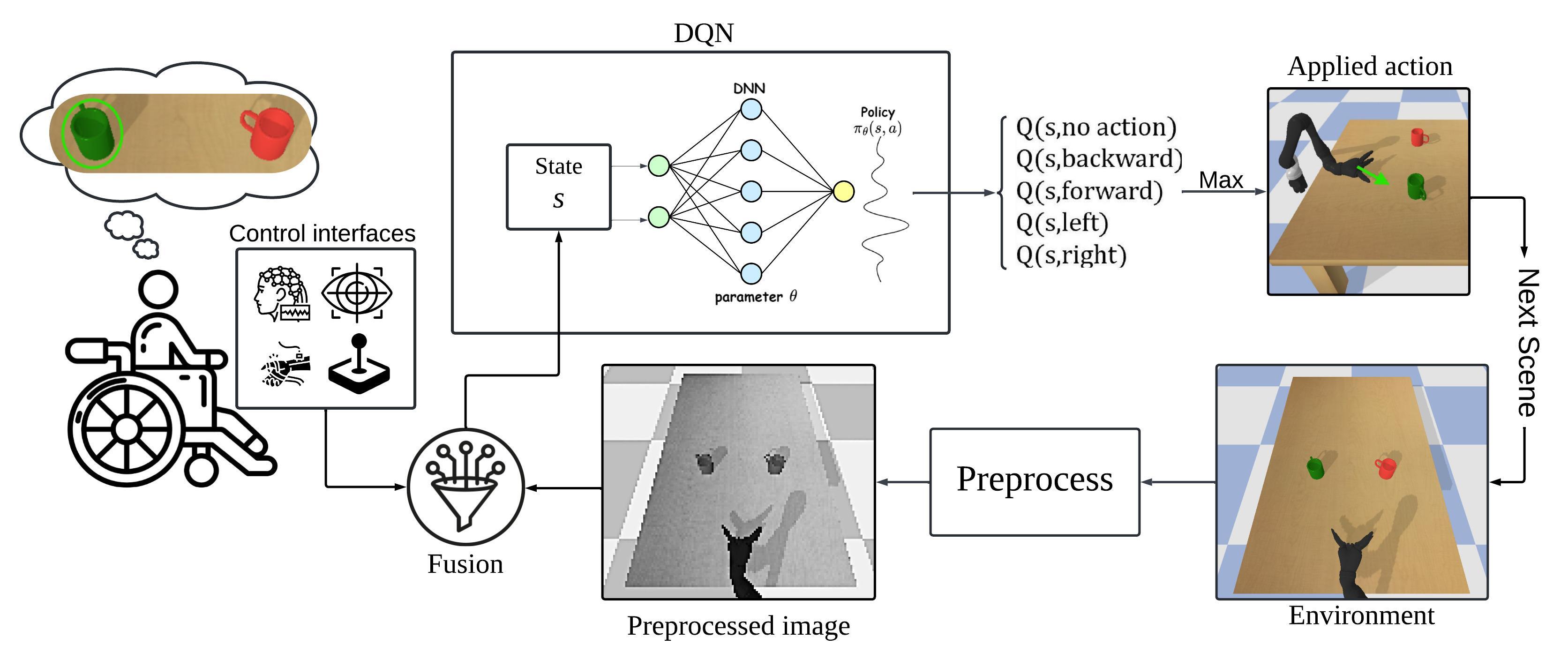}
\caption{Pipeline of the STREAMS framework for adaptive control of an assistive robotic arm using multimodal inputs coming from potential control interfaces and environment perception.}
\label{block_diagram}
\end{figure*}

We propose STREAMS framework based on Deep Q-Networks (DQN) for the control problem defined as a Partially Observable Markov Decision Process (POMDP). The main goal of our framework is to enable the end-effector to reach and grasp the intended object following a smooth trajectory based on environmental perception even with imprecise human input. As illustrated in Fig. \ref{block_diagram}, the STREAMS framework captures and preprocesses the environment image, and analyzes it with user input coming from the potential control interface to determine optimal robotic arm actions using a Deep Q-Network. After executing the chosen action, the resulting new state of the environment is fed back into the system, creating an adaptive control loop.

\begin{figure*}[ht!] 
\centering
\includegraphics[width=6.5in]{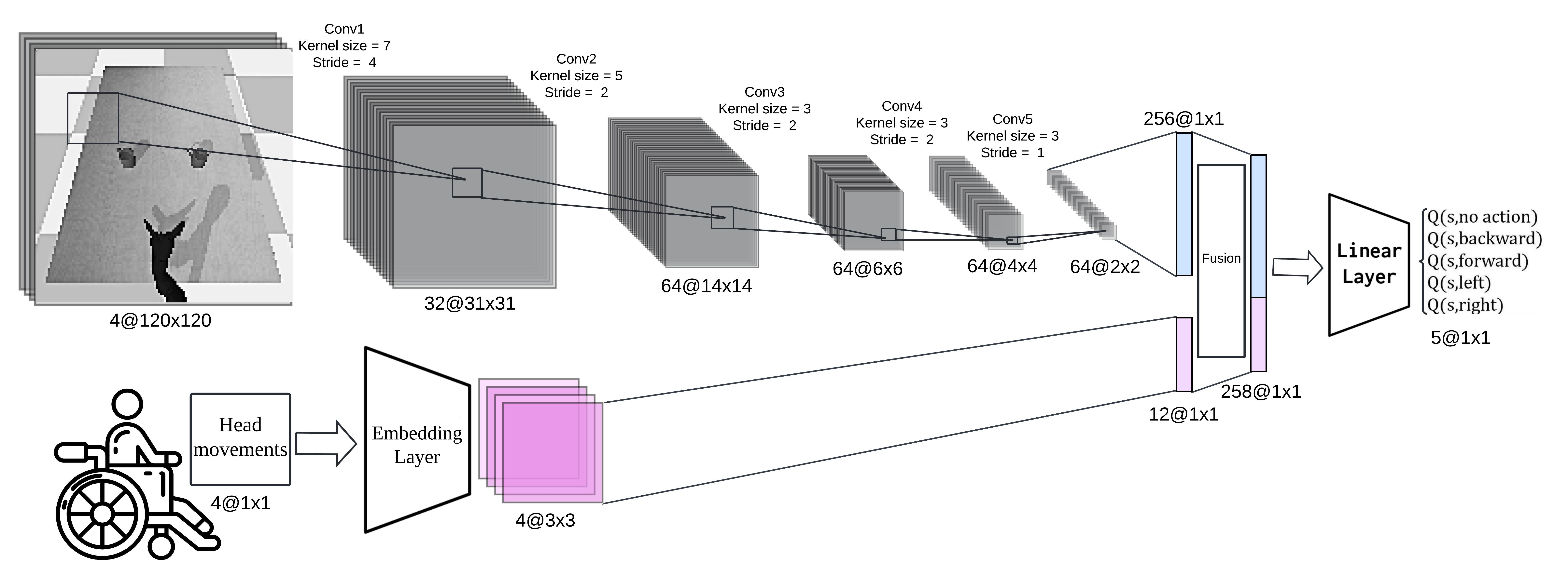}
\caption{The STREAMS framework's Deep Q-Network architecture processes stacked grayscale images and user inputs from the last four timesteps. Grayscale inputs pass through five convolutional layers, while user inputs are handled via an embedding layer, with both fused to output Q-values for five possible actions.}
\label{architecture}
\end{figure*}

Let \( \mathcal{S} \) denote the state space, which is composed of two key components: environment perception and human input. Assuming a severely impaired user (tetraplegia), we used head movement input with injected noise to simulate the imprecision typically seen in biosignal decoders. The environment perception is captured through a sequence of 2D grayscale images \( I_t \in \mathbb{R}^{H \times W} \), while the unreliable and imprecise human control input is represented as  \( \tilde{h_t}\). The state \( s_t \) at time \( t \) is defined as the concatenation of the last \( N \) frames:

\begin{equation}
    s_t = \{I_{t-N+1}, \ldots, I_t, \tilde{h}_{t-N+1}, \ldots,\tilde{h_t}\}
\end{equation}

In our case study, we consider \( \tilde{h_t} \in \{-1, 0, 1\} \), which represents decoded values of head movements corresponding to left, neutral, and right, respectively. The action space \( \mathcal{A} \) consists of five possible actions: forward, backward, right, left, and hold. We focus on controlling the end-effector position directly, while the positions of other joints in the robotic arm are determined using inverse kinematics. The Q-value function \( Q(s_t, a) \) is approximated by a DQN, which integrates convolutional layers to process image inputs, embedding layers to represent human inputs, and a late fusion mechanism to combine information from both modalities. The output of the fusion mechanism is passed through linear layers to make the final action decision as illustrated in Fig \ref{architecture}.


We explore three distinct reward functions to guide the learning process:

1. \textit{Binary reward} (\( R_1 \)): A simple binary reward where success is defined by the end-effector reaching and grasping the target object.

\begin{equation}
    R_1(s_t, a_t) = 
\begin{cases} 
1, & \text{if success} \\
0, & \text{otherwise}
\end{cases}
\end{equation}

2. \textit{Distance-based reward} (\( R_2 \)): This reward also incorporates the distance between the end-effector and the target object, rewarding the agent for reducing this distance.

\begin{equation}
    R_2(s_t, a_t) = 
\begin{cases} 
1, & \text{if success} \\
\alpha \cdot \left(1 - \frac{d_t}{d_{\text{init}}}\right), & \text{if episode is ongoing}\\
0, & \text{otherwise}
\end{cases}
\end{equation}

3. \textit{Time-penalized distance-based reward} (\( R_3 \)): This reward is similar to the distance-based reward but includes a time penalty to encourage faster task completion.

\begin{equation}
    R_3(s_t, a_t) = R_2(s_t, a_t) - \beta \cdot t,
\end{equation}

In these functions, \(d_t\) is the distance between end-effector and the target at each time step, \(d_{init}\) is the initial distance at the beginning of each episode, and \(\alpha\) and \(\beta\) are hyperparameters between 0 and 1.  A threshold is set for the euclidean distance between the end-effector and the target. When this distance exceeds the threshold, a grasp action is automatically triggered. The episode then concludes with the grasping attempt being evaluated as either successful or failed. Additionally, if the robot does not reach the target within 18 time steps, the episode is deemed a failure. 

As mentioned before, biosignal-based control systems, rely on discrete classification outputs from machine learning decoders to translate user intent into robotic actions. However, decoder outputs are inherently prone to misclassification due to signal variability and noise, leading to jerky and unstable control trajectories in real-world applications. Different studies \cite{rahman2019four, zhao2022deep, ai2019feature, ge2014classification} consistently show that for the multiclass motor imagery classification problem using EEG, which is among the most challenging biosignals to classify accurately, typical decoder accuracies range between 70\% and 90\%, indicating that misclassification rates can be as high as 30\% in practical applications.  To simulate real-world classification errors in biosignal-based control systems, we introduce a probabilistic flipping function, where ideal control inputs are randomly misclassified with probability \( p \) (Equation \ref{noise_equ}). This approach is inspired by common noise models used in EEG-based brain-computer interfaces and noisy reinforcement learning environments, where errors are modeled as stochastic perturbations to approximate real-world decoder inaccuracies \cite{ai2019feature, rahman2019four}. Unlike continuous noise models, such as Gaussian or uniform noise, the flipping function is more suitable for discrete control signals, such as categorical decisions in EEG, EMG, and IMU-based classification systems \cite{zhao2022deep, piastra2021comprehensive}.

During training, a synthetic user is developed to simulate the unreliable nature of human control input by generating ideal inputs (\(h^{ideal}\)) and adding noise to them. The ideal inputs represent the optimal, precise control inputs that guide the end-effector to the target object. This noise is introduced by randomly flipping input values with a probability (\(p\)), creating noisy and inaccurate inputs (\(\tilde{h_t}\)). By adjusting the noise level, input uncertainty can be varied. This synthetic approach eliminates the need for datasets and human intervention, speeding up the training process and improving the framework's ability to handle varying input quality. The input \(\tilde{h_t} \) is generated from the ideal input \(h^{ideal}_t\) as follows:

\begin{equation}
    \tilde{h_t} = 
\begin{cases} 
h^{ideal}_t, & \text{with probability } 1 - p \\
\text{Flip}(h^{ideal}_t), & \text{with probability } p 
\end{cases}
\label{noise_equ}
\end{equation}

Here, the function \( \text{Flip}(h^{ideal}_t) \) randomly changes \( h^{ideal}_t \) to one of the other two possible values \( \{-1, 0, 1\} \).  In our framework, we simulate three distinct levels of decoder noise by assigning probabilistic misclassification rates of \( p = 0.2 \), \( p = 0.3 \), and \( p = 0.4 \) during the training phase.

The pseudo-code for the training phase of the algorithm is presented in Algorithm \ref{alg:dqn}. The DQN in the STREAMS framework is trained using experience replay, where past experiences \( (s_t, a_t, r_t, s_{t+1}) \) are stored in a replay buffer, and mini-batches are sampled to update the network parameters \( \theta \). A target network with parameters \( \theta' \) is used to stabilize learning. The loss function involves the difference between the predicted and target Q-values, incorporating the reward \( r \) and discount factor \( \gamma \). The performance is evaluated on metrics such as task success rate, trajectory efficiency, and robustness to input noise. STREAMS is trained in a simulated environment with a hyperparameter random search using the Adam optimizer and an \(\epsilon\)-greedy strategy for exploration, with \( \epsilon \) decaying from 0.9 to 0.1. Training includes 25,000 episodes and randomization of cup positions to prevent overfitting and enhance generalization across tasks.

\begin{algorithm}[ht]
\caption{DQN-Based Reach-to-Grasp Control with Synthetic Noise}
\small
\begin{algorithmic}
\State Initialize replay buffer $\mathcal{D}$
\State Initialize primary network $Q(s, a; \theta)$ and target network $Q'(s', a'; \theta')$ with random weights
\For{each episode}
    \State Initialize state $s_1 = \{I_{1-N+1}, \ldots, I_1, \tilde{h}_{1-N+1}, \ldots, \tilde{h}_1\}$
    \For{each step $t$ in episode}
        \State With probability $\epsilon$, select random action $a_t$
        \State Otherwise, select $a_t = \arg\max_a Q(s_t, a; \theta)$
        \State Execute action $a_t$ 
        \State Observe reward $r_t$ and next state $s_{t+1}$
        \State Generate noisy human input $\tilde{h}_t$:
        \If {With probability $1 - p$}
            \State set $\tilde{h}_t = h^{ideal}_t$
        \Else
            \State set $\tilde{h}_t = \text{Flip}(h^{ideal}_t)$
        \EndIf
        \State Store transition $(s_t, a_t, r_t, s_{t+1})$ 
        \State Sample mini-batch of transitions $(s_j, a_j, r_j, s_{j+1})$ 
        \State Compute target $y_j = r_j + \gamma \max_{a'} Q'(s_{j+1}, a'; \theta')$
        \State Perform gradient descent step on $(y_j - Q(s_j, a_j; \theta))^2$
        \If {step mod target update frequency == 0}
            \State Update target network: $\theta' \leftarrow \theta$
        \EndIf
        \State Update state $s_t \leftarrow s_{t+1}$
    \EndFor
\EndFor
\end{algorithmic}
\label{alg:dqn}
\end{algorithm}
\begin{figure}[ht!] 
\centering
\includegraphics[width=1.9 in]{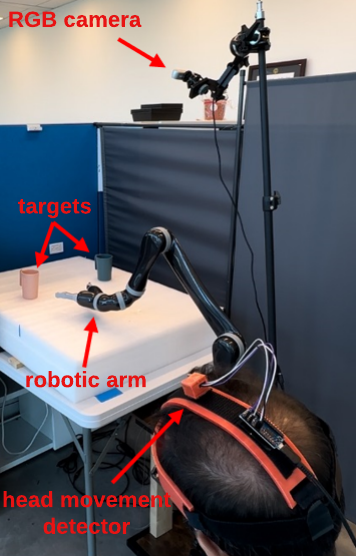}
\caption{Components of the STREAMS framework user study}
\label{user_study}
\end{figure}
\subsection{User study design}

This IRB-approved study involved five healthy participants aged 22-36 years. The experimental setup included a Kinova Jaco robotic arm, an RGB camera for workspace perception, an Inertial Measurement Unit (IMU)-based head movement detector, and two randomly selected target objects which are supposed to be grasped by user inputs (see Fig. \ref{user_study}). The head movement detector measured the angle of the head toward the sides and decoded these movements into three discrete control inputs: right, left, and neutral. These user inputs were used to guide the robotic arm to reach two different target cups. The framework was generalized to handle a wide range of target objects including different types of cups, bottles, and containers. The reach-to-grasp operation in a planar 2D space required comparing Manual (pure head movement control) and Assistive (with STREAMS) modes. Each participant completed 40 trials: 20 in Manual mode and 20 in Assistive mode, with 10 trials for each target (left and right) in both modes. Target positions remained constant throughout the experiment to ensure a systematic user study across participants. In this case study, users controlled lateral motion through head movements, while the end-effector constantly moved forward with each step. Grasping was automatically triggered whenever the end-effector was in proximity to the target sufficiently.

Before data collection, participants were trained in both modes and completed practice trials. Mode order was randomized to minimize learning effects. Data collection included video recording, trajectory extraction, targets and end-effector position tracking, and success rate recording. Participants also completed questionnaires on convenience, effort, and satisfaction for each mode. Performance metrics compared trajectory qualities, success rates, and user feedback between Manual and Assistive modes. A paired t-test was used to determine the statistical significance of observed differences.


\section{Results \& Discussion}
\subsection{Simulation}
\begin{figure}[ht!] 
\centering
\includegraphics[width=2.9in]{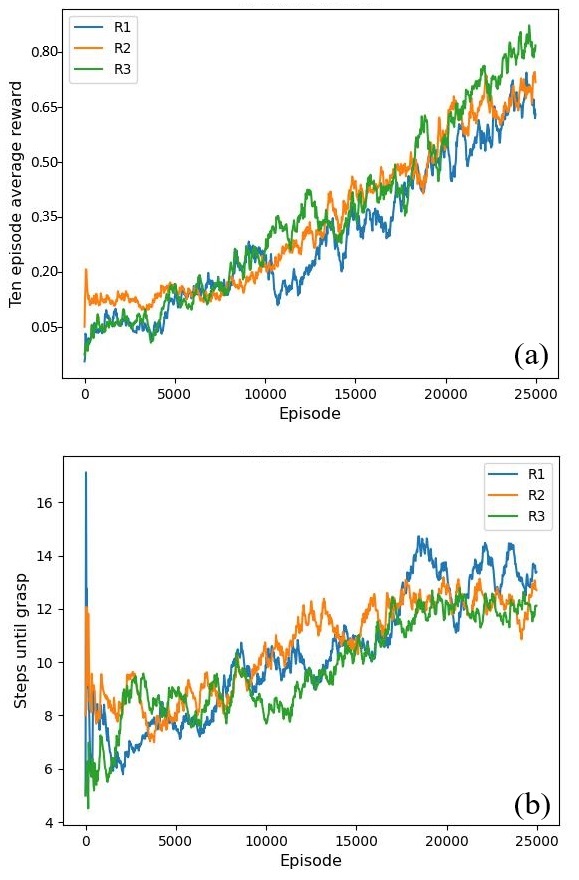}
\caption{Performance comparison of different reward functions (R1, R2, R3) in the STREAMS framework. (a) Ten-episode average reward over 25,000 episodes. (b) Number of steps required to grasp the target object per episode.}
\label{rewards_steps}
\end{figure}

In Fig. \ref{rewards_steps}, we investigated the performance of various reward functions designed to guide the robotic control system using the STREAMS framework. Three reward functions (R1, R2, and R3) were evaluated over 25,000 episodes, with R3 demonstrating better performance, reaching a peak average reward of 0.92. R1 and R2 also showed notable improvement, achieving maximum rewards of 0.71 and 0.73 respectively  (Fig. \ref{rewards_steps}a). All three functions exhibited an overall trend of improvement with increased training. In terms of efficiency, R3 achieved the quickest average grasping time at 11.90 steps, followed by R2 at 12.62 steps, and R1 at 13.54 steps  (Fig. \ref{rewards_steps}b). This shows that R3's time-penalized distance-based reward system was the most effective, while R2's distance-based rewards outperformed R1's binary reward system. The consistent learning curves across all three reward functions suggest that the STREAMS framework is robust and adaptable to different reward structures. 

\begin{table}[ht]
\centering
\caption{Success rates and standard deviations for manual and assistive modes at different noise levels.} 
\resizebox{1.03\linewidth}{!}{ 
\begin{tabular}{cccccc}
\toprule
\textbf{Noise level} & \textbf{Manual mode SR (\%)} & \multicolumn{3}{c}{\textbf{Assistive mode SR (\%)}} \\ \cmidrule(lr){3-5} 
& & \textbf{R1} & \textbf{R2} & \textbf{R3} \\ \midrule
\textbf{Low (0.2)} & 80.53 $\pm$ 2.34 & 98.53 $\pm$ 3.12 & 98.93 $\pm$ 1.87 & 99.33 $\pm$ 2.65 \\ 
\textbf{Medium (0.3)} & 74.26 $\pm$ 1.98 & 96.26 $\pm$ 4.11 & 97.83 $\pm$ 2.76 & 98.26 $\pm$ 3.44 \\ 
\textbf{High (0.4)} & 68.90 $\pm$ 3.75 & 91.26 $\pm$ 4.02 & 93.43 $\pm$ 2.49 & 95.26 $\pm$ 1.91 \\ 
\bottomrule
\end{tabular}
}
\label{tab:success_rates}
\end{table}

\begin{figure*}[ht!] 
\centering
\includegraphics[width=6.5in]{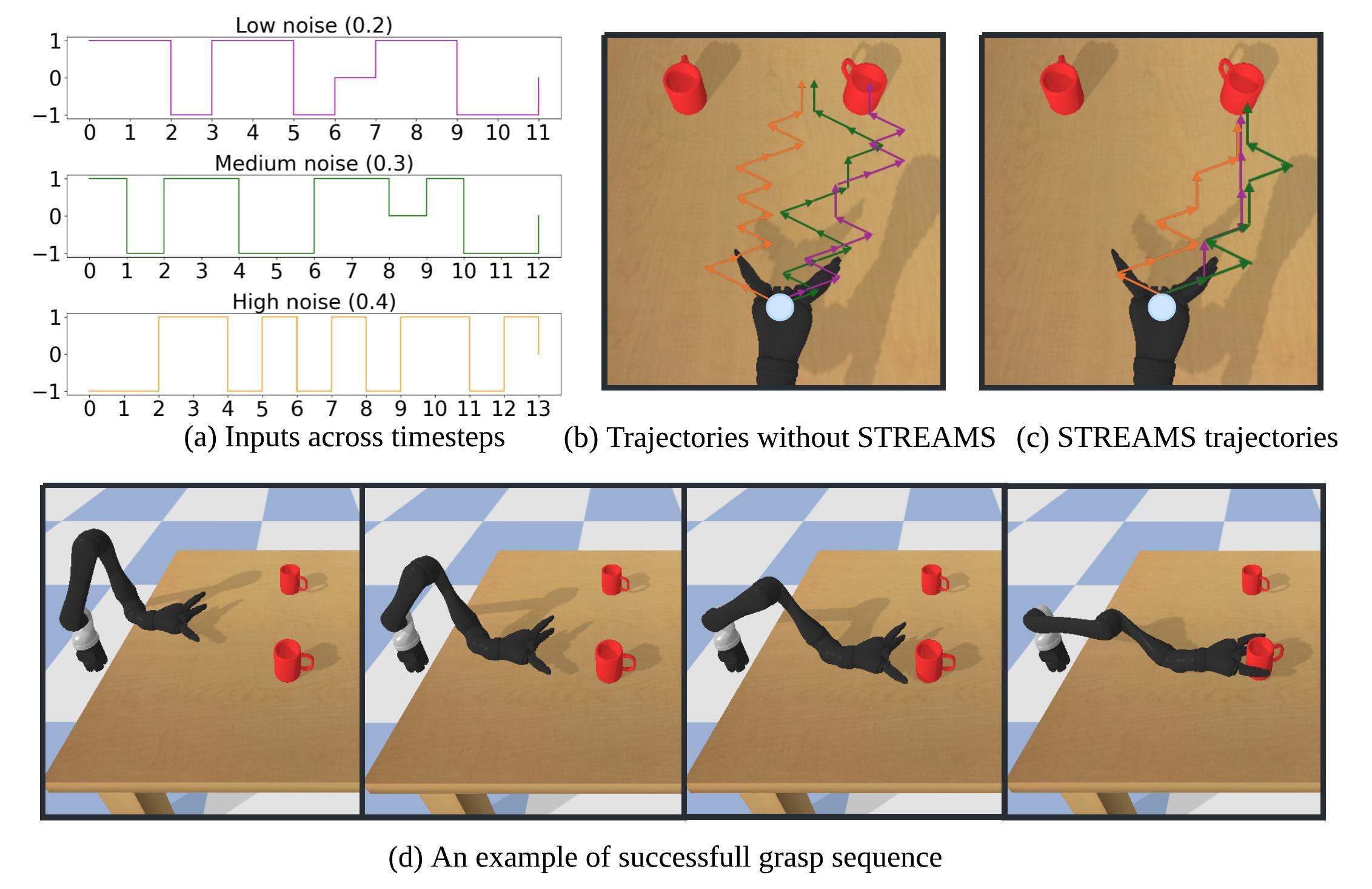}
\caption{Comparison of manual and STREAMS-assisted control under varying noise levels in the simulation environment. (a) Input signals. (b) Resulting trajectories in manual mode. (c) Improved trajectories using STREAMS in assistive mode. (d) Sequential snapshots of successful object grasping using STREAMS in a 3D environment.}
\label{sim_trajectories}
\end{figure*}

In Table \ref{tab:success_rates}, we presented a comparison of success rates (SR) between manual and assistive modes across various noise levels and reward functions over 3000 test episodes. The results showed the assistive mode consistently outperformed the manual mode across all noise levels, with success rates above 91\% even at the highest noise level. R3 achieved the best results, reaching 99.33\% success under low noise (0.2) and maintaining 95.26\% at high noise. STREAMS also demonstrated resilience to noise, with a minor performance drop compared to the sharper decline in manual mode, highlighting its robustness in noisy environments.

In Fig. \ref{sim_trajectories}, we illustrated the impact of noise on input signals and the resulting trajectories in both manual and assistive modes. Fig. 6(a) displays the input signals for low (purple), medium (green), and high (orange) noise levels. As the noise level increases, we can observe more frequent and erratic changes in the input signal. Fig. 6(b) and 6(c) show the traveled trajectories in a simulated environment. Fig. 6(b) represents the manual mode, where the trajectories are visibly erratic and inefficient, particularly for the medium and high noise levels. These paths demonstrated significant deviation and unnecessary movements, reflecting the difficulty of precise control with direct noisy inputs. In contrast, Fig. 6(c) shows the trajectories generated by the STREAMS framework in assistive mode. The paths are noticeably smoother and more direct across all noise levels. Even with high noise, the STREAMS-assisted trajectory maintains a relatively straight path toward the target object. Fig. 6(d)  shows a series of snapshots illustrating an example of a successful attempt where the robot arm moved toward the target object and grasped it.

\subsection{User experiments}

\begin{table*}[ht]
\centering
\caption{Summary of control methods and strategies in multimodal shared control assistive robotics.} 
\resizebox{1.\linewidth}{!}{ 
\begin{tabular}{lclccccl}
\toprule
\textbf{Study}             & \textbf{Approach} & \textbf{Modality} & \textbf{target-based} & \textbf{DoF} & \textbf{Dataset Req.} & \textbf{Accuracy} \\ \midrule
\textbf{Jain et al. 2015\cite{jain2015assistive}}   & Sequential policies & IMU + Vision & Yes & 6 (robot) & Yes & Not reported \\ 
\textbf{Downey et al. 2016 \cite{downey2016blending}} & Arbitrated policies & Spikes + Vision & No & 4 (3D + grasp) & Yes & 78\% \\ 
\textbf{Muelling et al. 2017\cite{muelling2017autonomy}} & Arbitrated policies & Spikes + Vision & No & 4 (3D + grasp) & Yes & 73.74\% \\ 
\textbf{Xu et al. 2019\cite{xu2019shared}}    & Arbitrated policies & EEG + Vision & Yes & 2 (planar) & Yes & ~70\% \\ 
\textbf{Beraldo et al. 2022\cite{beraldo2022shared}} & Fused policies & EEG + Vision & No & 2 (planar) & Yes & 79.94\% \\ 
\textbf{Ours (STREAMS)}              & End-to-end self-learning policy & IMU* + Vision & No & 3 (planar + grasp) & No & 83\% \\ 
\bottomrule
\end{tabular}
}
\label{tab:summary}
\begin{flushleft}
\centering
\small * A generalizable approach for other signals as well, not limited to IMU input only.
\end{flushleft}
\end{table*}

\begin{figure}[ht!] 
\centering
\includegraphics[width=2.7in]{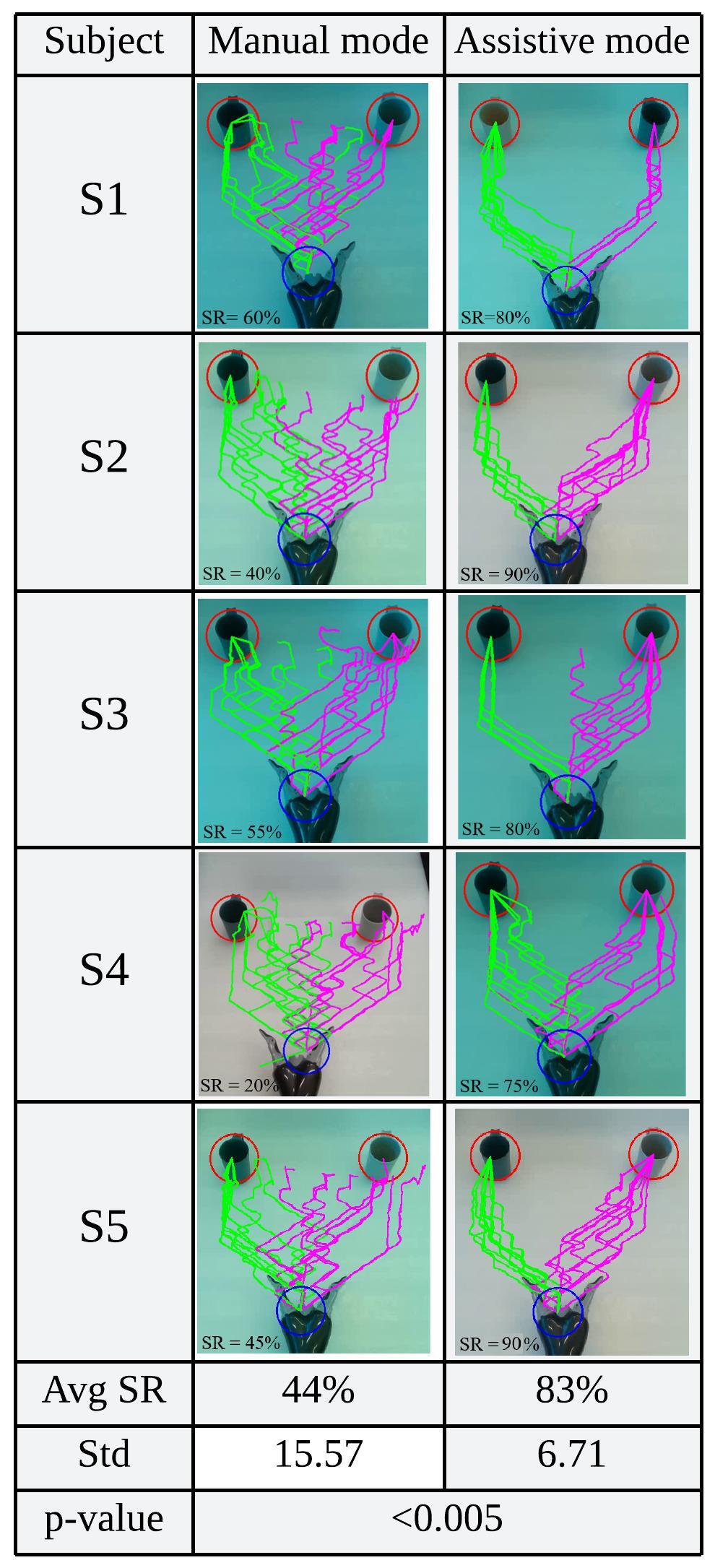}
\caption{Comparison of manual and STREAMS-assisted control modes across five subjects (S1-S5). Each row shows trajectories for left (green) and right (purple) target objects, with success rates (SR). The bottom rows present average success rates, standard deviations, and statistical significance (p-value) of the performance difference between modes.}
\label{user_study_trajectories}
\end{figure}

In Fig. \ref{user_study_trajectories}, we compared user performance in manual mode versus assistive mode using the STREAMS framework across five subjects (S1-S5). We employed medium-level input noise (0.3) for head movements. The best-performing model from our simulation was adopted for the assistive mode. The visual data clearly illustrates the improvements in control and efficiency offered by the assistive mode. In manual mode, highly erratic and inefficient trajectories were observed for all subjects. The green (left target) and purple (right target) paths showed considerable deviation, and unnecessary movements, and often failed to reach the intended targets. This was reflected in low success rates ranging from 20\% to 60\%, with an average of just 44\%. Conversely, the assistive mode displayed remarkably improved trajectories which consistently reached the intended targets. This improvement was quantified by substantially higher success rates, ranging from 75\% to 90\%, with an average of 83\%. The p-value of $<$0.005 indicated that the outperformance in assistive mode is statistically significant. 

The user study not only demonstrated quantitative improvements in performance but also revealed significant qualitative benefits of the STREAMS assistive mode. The questionnaire included questions on task load (Q1) and overall satisfaction (Q2) for each mode, both rated on a 1-5 scale (1 = low, 5 = high). For Q1, the manual mode average score was 4.0, while the assistive mode score was 2.2. This substantial difference indicates that users found the task much less demanding when using the assistive mode. The lower cognitive load in assistive mode suggests that STREAMS effectively reduces the mental and physical effort required to complete the grasping tasks. For Q2, the manual mode average rate was 1.8, while the assistive mode rate was 4.0. Users clearly preferred the STREAMS-assisted control, likely due to the increased success rate and reduced effort required. These subjective measures align well with the objective performance data shown in Fig. \ref{user_study_trajectories}. 

In Table \ref{tab:summary}, we presented a comparison of shared autonomy approaches in assistive robotics where STREAMS achieved the highest accuracy at 83\% with a 3-DoF system. Unlike other methods, STREAMS requires no pre-existing datasets or predefined arbitrated policies and it is highly adaptable to new environments and dynamic targets. Additionally, its flexible input modality (IMU*) supports various input types, extending beyond IMU and vision. While STREAMS has fewer degrees of freedom compared to some earlier studies \cite{jain2015assistive}, its non-target-based approach provides more flexibility in dynamic environments where targets may not be clearly defined. Moreover, STREAMS generates smooth and reliable trajectories even in the face of environmental uncertainties, dynamic targets, ambiguous human intentions, and imprecise user inputs, making it robust and reliable for real-world assistive applications.

\section{Conclusion}
This study introduced STREAMS as a task support framework for shared autonomy in assistive robotics, addressing the challenge of generating robust trajectories from unreliable and noisy user inputs for precise robotic grasping. The framework leverages deep reinforcement learning to refine control strategies dynamically, ensuring stable and smooth end-effector movements even under uncertain user input conditions.  Our generalizable framework integrated several key components, including an end-to-end self-learning policy that enables real-time interaction with dynamic user intent and adaptation to environmental changes. Unlike many existing approaches that rely on predefined datasets or explicit goal prediction, STREAMS utilizes a synthetic noise model to train its control policies, allowing it to adapt to different biosignal-based input modalities without requiring extensive data collection. This adaptability makes it particularly well-suited for assistive applications where user input variability is high. 

STREAMS demonstrated the ability to generate smooth and stable trajectories across various noise levels and user conditions, highlighting its robustness to decoder misclassifications and signal noise. The efficacy of the proposed framework was validated through extensive simulations and real-world user studies, showing significant improvements in task performance, user satisfaction, and control accuracy compared to manual control without task support. The framework consistently outperformed manual methods in both success rate and trajectory efficiency, reducing the cognitive and physical effort required from users to complete complex robotic manipulation tasks.  Future research directions could include expanding the framework’scapabilities to handle more complex environments and tasks.

\bibliographystyle{IEEEtran}
\bibliography{root}

\end{document}